\documentclass[conference]{IEEEtran}
\IEEEoverridecommandlockouts
\usepackage{cite}
\usepackage{amsmath,amssymb,amsfonts}
\usepackage{algorithmic}
\usepackage{graphicx}
\usepackage{textcomp}
\usepackage{xcolor}
\usepackage{soul}
\usepackage[utf8]{inputenc}
\usepackage[small]{caption}
\usepackage{url}  
\usepackage{katsuki}
\usepackage{array}
\usepackage{bm}
\usepackage[caption=false,font=footnotesize]{subfig}
\usepackage{stfloats}
\usepackage{booktabs}

\begin{document}
\title{Time-Discounting Convolution for\\Event Sequences with Ambiguous Timestamps}

\author{\IEEEauthorblockN{Takayuki Katsuki, Takayuki Osogami,\\Akira Koseki, Masaki Ono, Michiharu Kudo}
\IEEEauthorblockA{IBM Research - Tokyo\\
Tokyo, Japan\\
e-mail: \{kats,osogami,akoseki,moono,kudo\}@jp.ibm.com}
\and
\IEEEauthorblockN{Masaki Makino, Atsushi Suzuki}
\IEEEauthorblockA{Department of Endocrinology and Metabolism,\\Fujita Health University\\
Aichi, Japan\\
e-mail: \{makinom,aslapin\}@fujita-hu.ac.jp}
}

\maketitle

\begin{abstract}
This paper proposes a method for modeling event sequences with ambiguous timestamps, a time-discounting convolution. Unlike in ordinary time series, time intervals are not constant, small time-shifts have no significant effect, and inputting timestamps or time durations into a model is not effective. The criteria that we require for the modeling are providing robustness against time-shifts or timestamps uncertainty as well as maintaining the essential capabilities of time-series models, i.e., forgetting meaningless past information and handling infinite sequences. The proposed method handles them with a convolutional mechanism across time with specific parameterizations, which efficiently represents the event dependencies in a time-shift invariant manner while discounting the effect of past events, and a dynamic pooling mechanism, which provides robustness against the uncertainty in timestamps and enhances the time-discounting capability by dynamically changing the pooling window size. In our learning algorithm, the decaying and dynamic pooling mechanisms play critical roles in handling infinite and variable length sequences. Numerical experiments on real-world event sequences with ambiguous timestamps and ordinary time series demonstrated the advantages of our method.
\end{abstract}

\begin{IEEEkeywords}
Electronic health records, Electronic healthcare, Health information management, Event sequence, Convolutional neural networks, Time series analysis
\end{IEEEkeywords}
\renewcommand{\thefootnote}{\fnsymbol{footnote}}
\footnote[0]{\copyright 2018~IEEE~~DOI 10.1109/ICDM.2018.00139}
\renewcommand{\thefootnote}{\arabic{footnote}}

\section{Introduction}
\begin{figure}[tb]
	\centering
	\includegraphics[width=65mm]{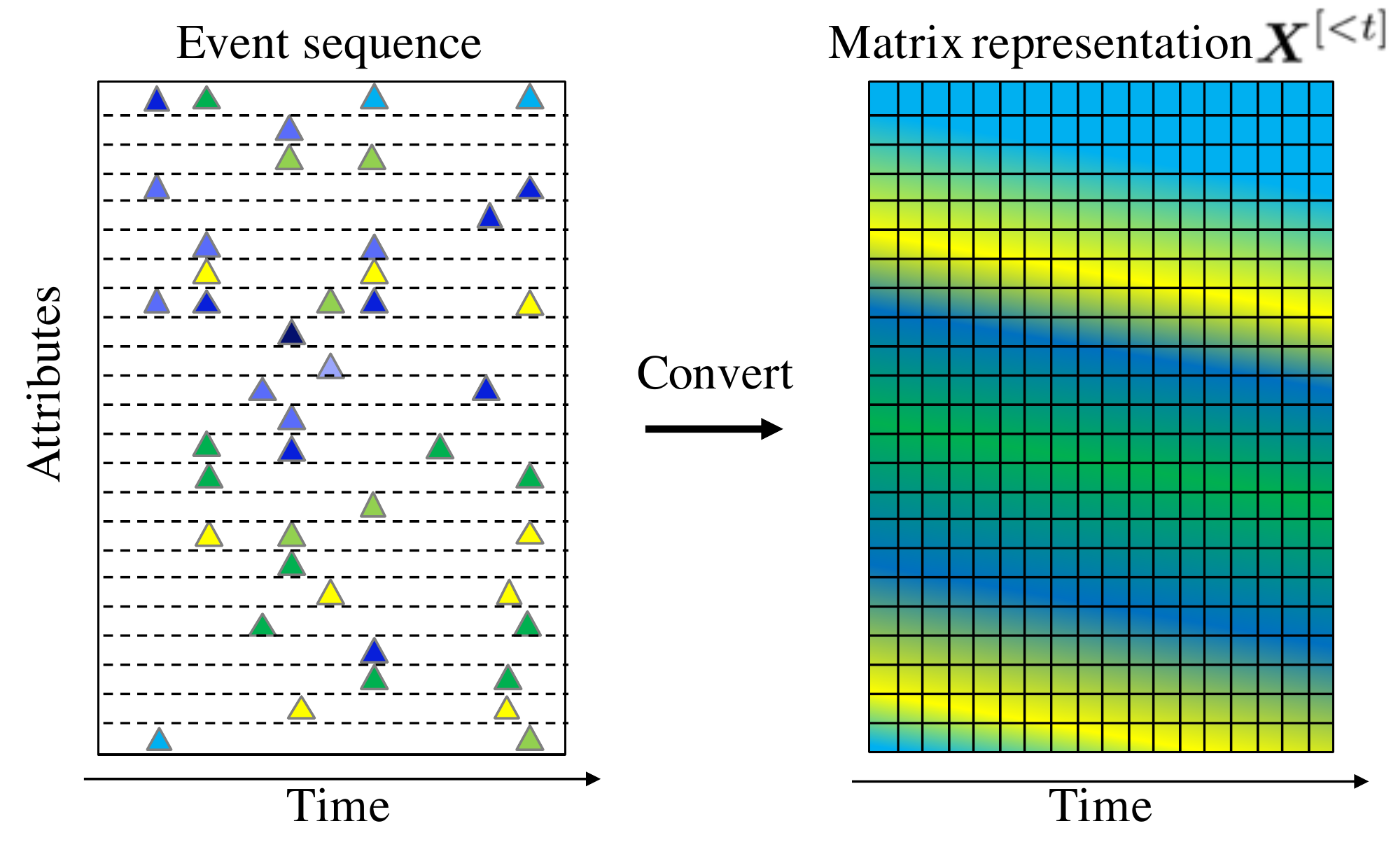}
    \vspace{-5pt}
	\caption{Example of event sequence with ambiguous timestamps and its matrix representation. Horizontal and vertical axes respectively correspond to timestamps and attributes. Triangles indicate attribute values of events at corresponding times; yellow and blue mean high and low values, respectively.}
	\label{FigMatrix}
\end{figure}

Temporal event sequences record timestamped events, which are ubiquitous in the real world and are addressed in various data mining problems. We address the scenario where the attached timestamps are ambiguous, as shown in Fig.~\ref{FigMatrix}. Such sequences are found in events that are recorded passively, i.e., without observers' instantaneous control. The timestamps do not represent when events have occurred but represent when they were recorded. In such cases, the time intervals are variable and the timestamps are not reliable. Thus, small time shifts in the observed sequences have no significant effect, and there are uncertainties in recorded timestamps.
In medical informatics, for example, attention is being paid to analyzing such passively recorded data, as found in electronic health records (EHRs) for predicting risks to patients~\cite{cheng2016risk,che2017boosting,katsuki2018risk}. They are recorded when a patient is treated at a hospital.

Modeling event sequences is one of the fundamental problems in data mining, such as analyses on sensor time-series, economic data, and EHRs. While we can assume independent observations for non-sequential data, we must take into account the dependencies between successive events.
In standard modeling approaches, one of two major assumptions is made: the observations occur at regular intervals (so the timestamps can be ignored) or at irregular intervals (so the timestamps must be considered).
The first approach is typically used for time-series and language modeling. Once the order of the sequence is incorporated into the model, exact timestamps and durations are unimportant.
There are many established models, including vector autoregressive (VAR) models~\cite{lutkepohl2005new}, hidden Markov models~\cite{baum1966statistical}, recurrent neural networks (RNN)~\cite{rumelhart1985learning}, long short-term memory (LSTM) models~\cite{hochreiter1997long}, and Boltzmann machines for time series~\cite{taylor2007modeling,hinton2000spiking,sutskever2007learning,sutskever2009recurrent,DyBM,osogami2015learning,NonlinearDyBM,FunctionalDyBM,BidirectionalDyBM}. If the time intervals are not constant, the performance is degraded in return for simpler modeling for temporal dependencies.

The second approach is typically used for asynchronously observed event sequences, such as in log records and process-series data.
Along with other features representing events, the timestamps or intervals between events are explicitly input to the model for encoding the dependencies between successive events.
RNN models have been used for this purpose~\cite{choi2016doctor,xiao2017modeling,che2018recurrent}. However, if the timestamps are not reliable, directly inputting them into the model might not be effective.

For modeling the event sequences with ambiguous timestamps, we addressed three major modeling requirements.
\begin{LaTeXdescription}
    \item[Time-shift invariance]~\\
    The model should be invariant against time shifts, since almost the same patterns would be recorded with a time shift if there is no instantaneous and on-demand control.
    \item[Robustness against timestamp uncertainty]~\\
    The model should be robust against uncertainty in timestamps, since the timestamps represent when they were recorded but not when events have occurred.
    \item[Natural forgetting]~\\
    The ability to forget meaningless past information should be inherited from time-series models, enabling the handling of infinite sequences and long-term dependency.
\end{LaTeXdescription}

We propose a {\it time-discounting convolution} method that uses a specific convolutional structure and a {\it dynamic pooling} mechanism. Our convolutional structure has a uni-directional convolution mechanism across time with two kinds of parameter sharing for efficiently representing the dependency between events in a time-shift invariant manner. It also has a mechanism of naturally forgetting by discounting the effects of past observations. The structure is based on the eligibility trace in dynamic Boltzmann machines (DyBMs)~\cite{DyBM,osogami2015learning,NonlinearDyBM,FunctionalDyBM,BidirectionalDyBM}, whose learning rules have a biologically important characteristic, i.e., spike-timing-dependent synaptic plasticity (STDP). This is our first contribution. The dynamic pooling mechanism provides robustness against the uncertainty in timestamps and enhances the time-discounting capability by dynamically changing the window size in its pooling operations. This is our second contribution.

Several time-convolutional models have been proposed to capture shift-invariant patterns while representing temporal dependencies. The time-delay neural network~\cite{waibel1989phoneme} is a pioneering model. Convolutional neural networks (CNNs) have recently been applied to sequential data, such as stock price movements~\cite{ding2015deep}, sensor outputs~\cite{singh2017convolutional}, radio communication signals~\cite{o2016unsupervised}, videos~\cite{bascol2016unsupervised}, and EHRs~\cite{cheng2016risk,katsuki2018risk,katsuki2018feature}. However, these models do not have a temporal nature, i.e., the natural forgetting capability, which is one of our modeling requirements. To represent the both convolutional and temporal natures, stacking combinations of convolutional models and time-series models have also been studied recently~\cite{sennrich2016neural,zheng2015conditional}. Our model inherently has both a time-series and convolutional aspects. This reduces the number of model parameters, which is quite useful when the amount of training data is limited.

We empirically evaluated the effectiveness of the proposed method in numerical experiments to examine its utility of the method for the real event sequences with ambiguous timestamps and general workability of the method for the real time-series data. We found that the proposed method improves predictive accuracy.

\section{Prediction from Temporal Event Sequences}
Our goal is to construct a model for predicting objective variables, $\by^{[t]}$ at time $t$, from an event sequence before time $t$, $\bX^{[<t]}$, where vector $\by^{[t]}$ can be a future event itself (autoregression) or any other unobservable variable (regression or classification), and $\bX^{[<t]}$ represents all the observed sequences before time $t$.

An event sequence is a set of records of events with timestamps. A record has the values of $D$ attributes, and each of the attributes may or may not be observed with other attributes at the same time, as shown in Fig.~\ref{FigMatrix}. For ease of analysis, the sequence $\bX^{[<t]}$  is usually represented as a $D \times T$ matrix~\cite{wang2012towards,katsuki2018risk,katsuki2018feature}. The horizontal dimension represents the timestamp at regular intervals with the highest temporal resolution in the sequence, and the vertical dimension represents the attribute values, as shown in Fig.~\ref{FigMatrix}, {\it i.e.}, $\bX^{[<t]} \equiv \{\bx^{[t-d]}\}_{d=t-T}^{t-1}$, where $\bx^{[t]}$ is the vector of attribute values at $t$.
If the event records are originally observed at regular intervals and all the attributes are always observed, the temporal event sequence is reduced to ordinary time-series data. If the original observation intervals vary over time and all the attributes are not observed simultaneously, several elements of the matrix will be missing. We replace missing values with large negative values (sufficiently lower than the minimum value of each attribute), which works well with our dynamic pooling described in the following section.

We learn the parameters $\btheta$ of our prediction model, $\bm{f}(\bullet,\btheta)$, by minimizing the objective function:
\begin{align}
    \label{EqPredictionError}
    \mathcal{L}(\btheta) &\equiv \sum_{t=1}^N \mathcal{L}^{[t]}(\btheta),~~~~\mathrm{where}\\
    \mathcal{L}^{[t]}(\btheta) &\equiv \mathrm{L}\big( \by^{[t]}, \bm{f}(\bX^{[<t]},\btheta)\big),\nonumber
\end{align}
where $N$ is the number of training samples ($N>T$) and $\mathrm{L}(\bullet)$ is a loss function, which is selected for each task and the corresponding objective variables $\by^{[t]}$, from mean squared error, cross entropy, log likelihood, and other functions.

By using the learned parameters, $\hat{\btheta}$, we can predict $\by$;
\begin{equation}
    \label{EqPrediction}
    \hat{\by}^{[t]} \equiv \bm{f} \big(\bX^{[<t]},\hat{\btheta}\big).
\end{equation}
We define our prediction model in the following section.

\section{Time-discounting Convolution}
\subsection{Prediction model with time-discounting convolution}
\begin{figure}[tb]
	\centering
	\includegraphics[width=85mm]{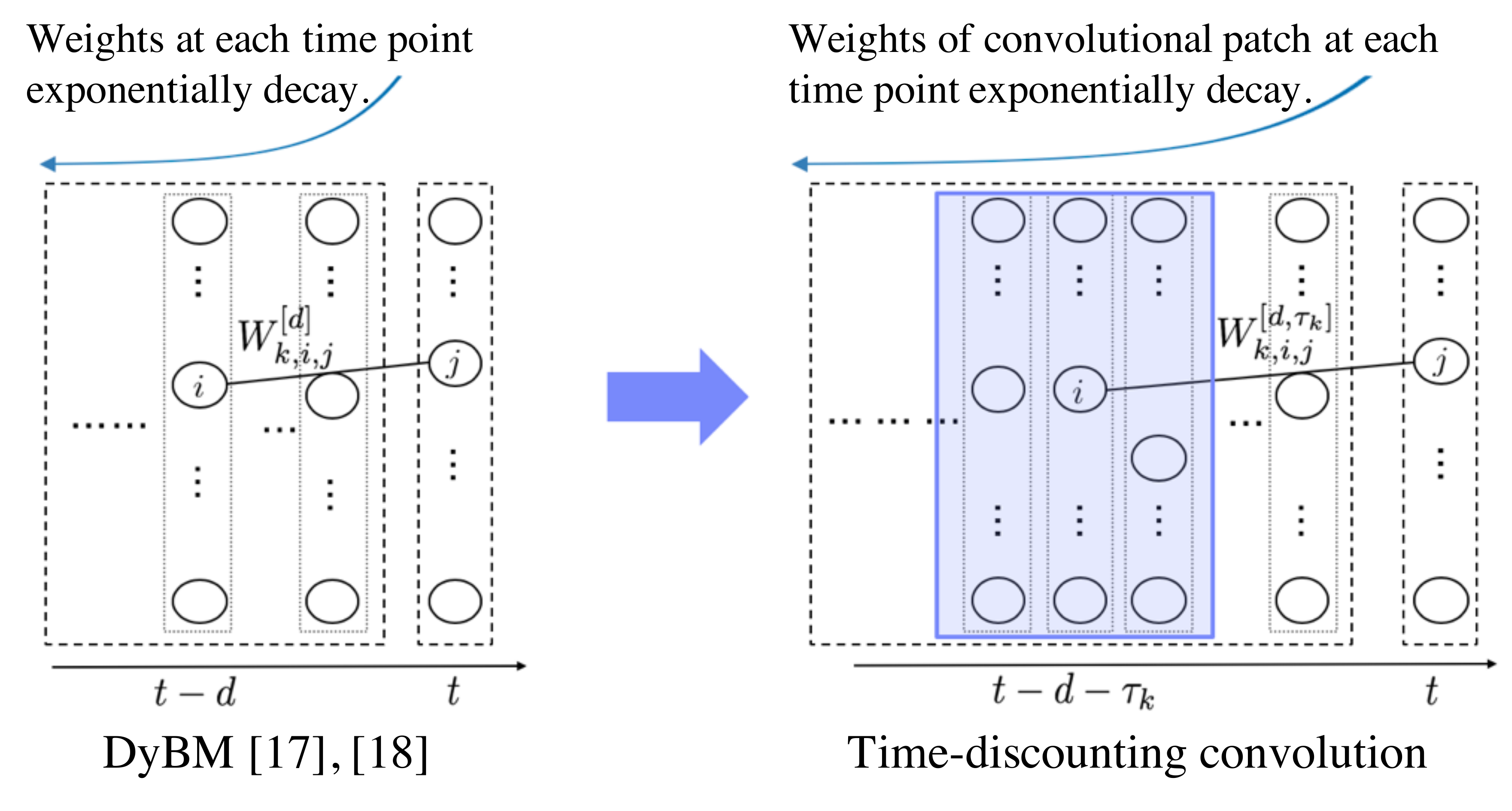}
	\caption{Proposed time-discounting convolution method.}
	\label{FigFramework}
\end{figure}
We propose a time-discounting convolution method for the prediction model. It has a convolutional structure across time, where the weight of a convolutional patch decays exponentially at each time point, as shown in Fig.~\ref{FigFramework}.
The proposed model has parameters $\btheta \equiv [\bW,\bb]$ and predicts the $j$-th element of $\by^{[t]}$ by the use of a non-linear function $h(\bullet)$ that maps a $K \times T$-dimensional input to a $1$-dimensional output:
\begin{align}
    \label{predict_CDyBM}
    &[f(\bX^{[<t]},\btheta)]_j\\
    &= h\bigg(\bigg\{\sum_{i} \sum_{\tau_k=0}^{T_k} x^{[t-d-\tau_k]}_i W^{[d,\tau_k]}_{k,i,j} - b_{k}
    \bigg\}_{1\le k \le K, 1\le d \le T}\bigg),\nonumber
\end{align}
where $W^{[d,\tau_k]}_{k,i,j}$ is a convolutional parameter across time with $\tau_k$ for the $i$-th attribute value $x_i^{[t-d]}$ at time $t-d$, and $T_k$ is the time length of the $k$-th convolutional patch. We use bias parameter $b_k$ individually for each of the $K$ feature maps. The non-linearity in $h(\bullet)$ makes this apparently redundant formulation meaningful, analogous to CNNs. We define specific functional forms of $h(\bullet)$ for each task in Section~\ref{sec:exp} along with the details of the implementation.

We use two different parametric forms for $W^{[d,\tau_k]}_{k,i,j}$:
\begin{align}
    \label{w_decay}
    W^{[d,\tau_k]}_{k,i,j} & = \lambda^{d+\tau_k} U_{k,i,j},~~\mathrm{and}\\
    \label{w_conv}
    W^{[d,\tau_k]}_{k,i,j} & = \mu^{d} V_{k,\tau_k,i,j},
\end{align}
where $\lambda, \mu \in [0,1)$ is the decay rate. Note that $W^{[d,\tau_k]}_{k,i,j}$ forms a tensor and corresponds to a patch in CNNs. Eq.~\eqref{w_decay} uses $U_{k,i,j}$ consisting of a single parameter across time for each $k$, $i$, and $j$. In the convolutional patch based on Eq.~\eqref{w_decay}, $U_{k,i,j}$ is replicated and used for multiple temporal positions of the time convolution with the decay rate of $\lambda^{d+\tau_k}$. The parameterization with shared parameters in Eq.~\eqref{w_decay} works similarly to the eligibility trace of DyBM, as shown on the left side of Fig.~\ref{FigFramework}; that is, the convolutional patch extracts the feature that represents the frequency of each observation and its distance from the current prediction.
Eq.~\eqref{w_conv} uses $V_{k,\tau_k,i,j}$ consisting of individual parameters for each time $\tau_k$ and other indexes $k$, $i$, and $j$. The $V_{k,\tau_k,i,j}$ forms the convolutional patch by itself and all the parameters in this patch decay together by $\mu^{d}$ in accordance with the timestamp $d$. We use Eqs.~\eqref{w_decay} and~\eqref{w_conv} in the same proportion in our $K$ feature maps.
The proposed method can capture discriminative temporal information in a shift-invariant manner because of its convolutional operation. Also, it can naturally forget past information, and prediction and gradients in learning do not diverge in the limit of $T \to \infty$ thanks to the decay rates $\lambda$ and $\mu$ in Eqs.~\eqref{w_decay} and~\eqref{w_conv}.

We can use our model as a layer in a neural network, can incorporate a neural network into our model via $h(\bullet)$, or as a preprocessor of the input $\bX^{[<t]}$ in Eq.~\eqref{predict_CDyBM}. In our experiments, we actually used our model along with a fully-connected layer and activation functions for prediction.
\subsection{Dynamic pooling}
\begin{figure}[tb]
	\centering
	\includegraphics[width=75mm]{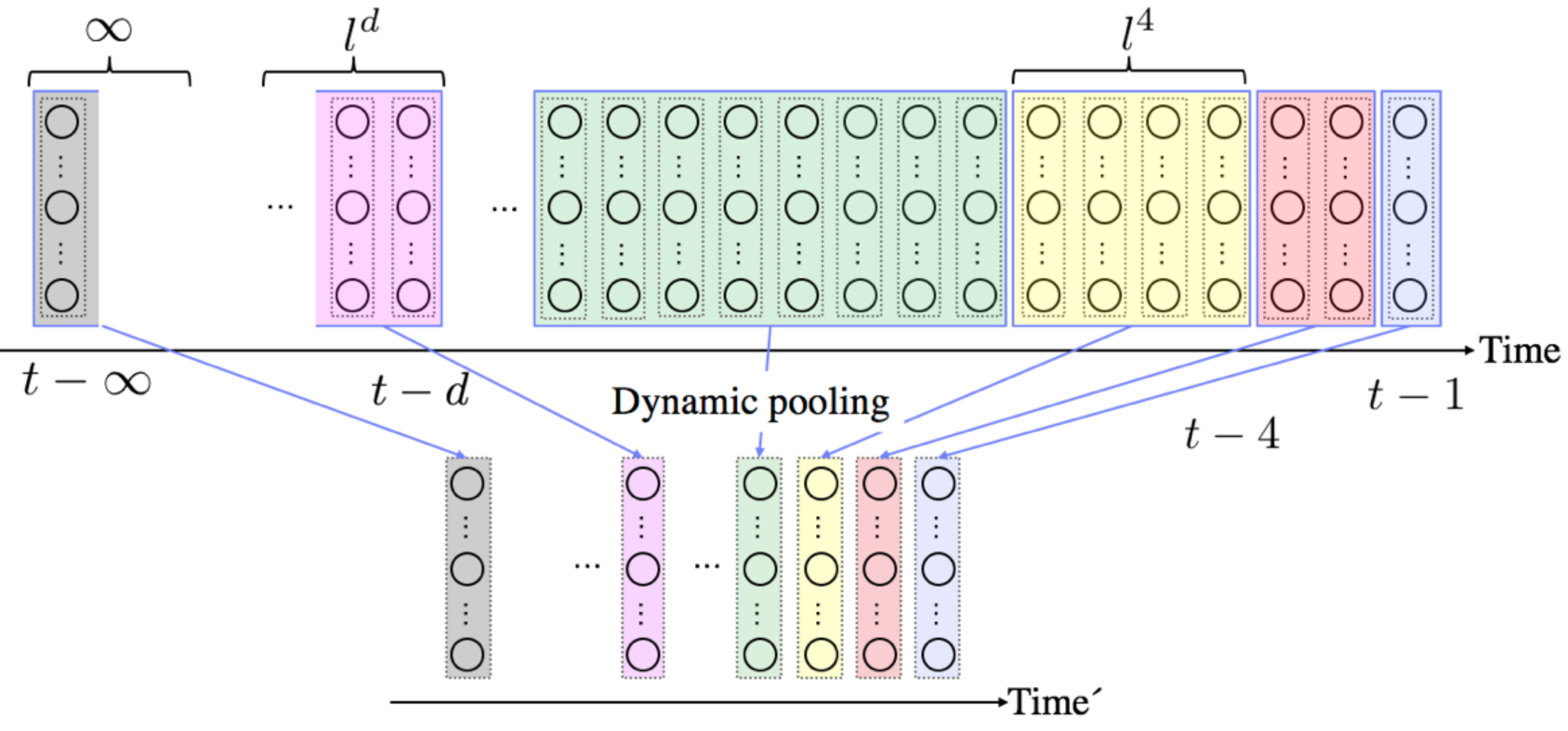}
	\caption{Dynamic pooling}
	\label{FigDynamicPooling}
\end{figure}
We introduce dynamic pooling as a powerful mechanism to avoid overfitting to vague timestamps and past meaningless information. Our pooling window increases in accordance with the time from prediction point $t$ as shown in Fig.~\ref{FigDynamicPooling}. Specifically, we let
\begin{equation}
    \label{dynamic_pooling}
    z^{[t]}_{i} = \max_{\tau_k \in [0,l_0\times l^t]} \big(x^{[t-\tau_k]}_{i}\big),
\end{equation}
where $l_0$ is the initial window size and $l$ is the growth rate of the window.
We can dynamically downsample the observed event sequences or latent representations by taking the maximum value over sub-temporal regions along with increasing the window size exponentially.

Dynamic pooling is used in the proposed method both as a preprocessor of input $\bX^{[<t]}$ and function $h(\bullet)$ in Eq.~\eqref{predict_CDyBM}. As a preprocessor of $\bX^{[<t]}$, we first apply dynamic pooling to a raw sequence then use the preprocessed sequence as $\bX^{[<t]}$. As $h(\bullet)$, we apply dynamic pooling to the latent representations that are the inputs to $h(\bullet)$.

Dynamic pooling leads to tractable analysis of the missing values by ignoring them in its max operation as the first pooling layer. It also enables us to easily handle infinite sequences when we make the final window size infinite. Also, we can handle the varying (horizontal) dimensions of $\bX^{[<t]}$ across different $t$ in the same manner as handling infinite sequences. The pooling layer after convolution works as an ordinary pooling method, i.e., the patterns having the largest effect are extracted. Because the rate of selecting each time point in the max operation decreases due to the growth of the window width in accordance with the time length from the prediction point $t$, the expected effect of each time point decays exponentially. This is also similar to the eligibility trace in DyBM.
We can define other pooling mechanisms, such as {\it dynamic mean-pooling}, by replacing the $\max(\bullet)$ operation with another operation.

\subsection{Learning Model Parameters}
In our experiments, we tackled autoregression and classification problems. We define the objective function $\mathcal{L}^{[t]}(\btheta)$ for each problem setting and derive the learning rules. For the autoregression problems, we use the L2-norm in Eq.~\eqref{EqPredictionError}:
\begin{equation}
    \label{EqPredictionError_autoreg}
    \mathcal{L}^{[t]}(\btheta) \equiv \big\| \bX^{[t]} - f(\bX^{[<t]},\btheta) \big\|^2,
\end{equation}
For the classification problems, we use the following form in Eq.~\eqref{EqPredictionError}:
\begin{equation}
    \label{EqPredictionError_class}
    \mathcal{L}^{[t]}(\btheta) \equiv \mathrm{L_c}\big( \by^{[t]}, \bsigma(\bm{f}(\bX^{[<t]},\btheta))\big),
\end{equation}
where the function $\mathrm{L_c}(\bullet)$ is the cross entropy, and function $\bsigma(\bullet)$ is the softmax function. For the classification problems, we use $\bsigma(f(\bX^{[<t]},\btheta)$ as the prediction model.

The model parameters are learned by mini-batch gradient descent. The update rule with $t$ is defined as
\begin{equation}
    \label{mini_batch_gradient_descent}
    \btheta \gets \btheta - \bmeta \frac{1}{M} \sum_{m=t-M+1}^{t}\frac{\partial \mathcal{L}^{[m]}(\btheta)}{\partial \btheta},
\end{equation}
where $\eta$ is the learning rate and $M$ is the mini-batch size. The specific gradients of the parameters are omitted due to space limitations (see the Appendix).
In the training phase, the dynamic pooling is simply passed through the gradient to the unit selected as maximum, analogous to ordinary max-pooling.
In the mini-batch gradient descent, the learning rate $\eta$ is controlled using the Adam optimizer with the hyperparameters recommended in~\cite{kingma2014adam}, and the mini-batches are set as $M=16$ examples. We used the same procedure for all the models we compared in our experiments.
The detailed settings of the hyperparameters $K$, $T_k$, $\lambda$, $\mu$, $l_0$, and $l$ are described for each experimental task in Section~\ref{sec:exp}.

\subsection{Relationships to Other Models}
Our model can be seen as a generalization of DyBM. The corresponding prediction model by the DyBM is defined as
\begin{align}
    \label{predict_DyBM}
    &[f_{\mathrm{DyBM}}(\bX^{[<t]},\btheta)]_j\\
    &= h\Big(\Big\{ \sum_{i} x^{[t-d]}_i W^{[d]}_{k,i,j} - \sum_{i}b_{k,i}\Big
    \}_{ 1\le k \le K, 1\le d \le T} \Big).\nonumber
\end{align}
This is a special case of our model (Eq.~\eqref{predict_CDyBM}) when $T_k=0$ for any $k$. In Eq.~\eqref{predict_DyBM}, the conduction delay of DyBM is assumed to be zero. In other words, we extend the summation and definition of the eligibility trace in DyBM to the $1D$-convolutional operation in our model. We also extended DyBMs to be applicable to classification problems and neural-network layers.

Our model reduces to a VAR model with lag $T$ if we use only Eq.~\eqref{w_conv}, let $h(\bullet)$ be the summation over $d$, and set $T_k=T$, $\mu=1$, and $k = 1$. Eq.~\eqref{predict_CDyBM} then reduces to
\begin{equation}
    \label{predict_VAR}
    [f_{\mathrm{VAR}}(\bX^{[<t]},\btheta)]_j = \sum_{i} \sum_{\tau=1}^{T} x^{[t-\tau]}_i V_{\tau,i,j} - b,
\end{equation}
where we omit $k$ and $d$ because of the above assumptions.

From the above relationships, our model can be seen as an ensemble of convolutional terms (generalization of eligibility trace of DyBM) and VAR-like terms. The convolution with $T_k \neq \infty~\mathrm{or}~T$ particularly differentiates our model from them.

\begin{figure}[tb]
	\centering
	\includegraphics[width=75mm]{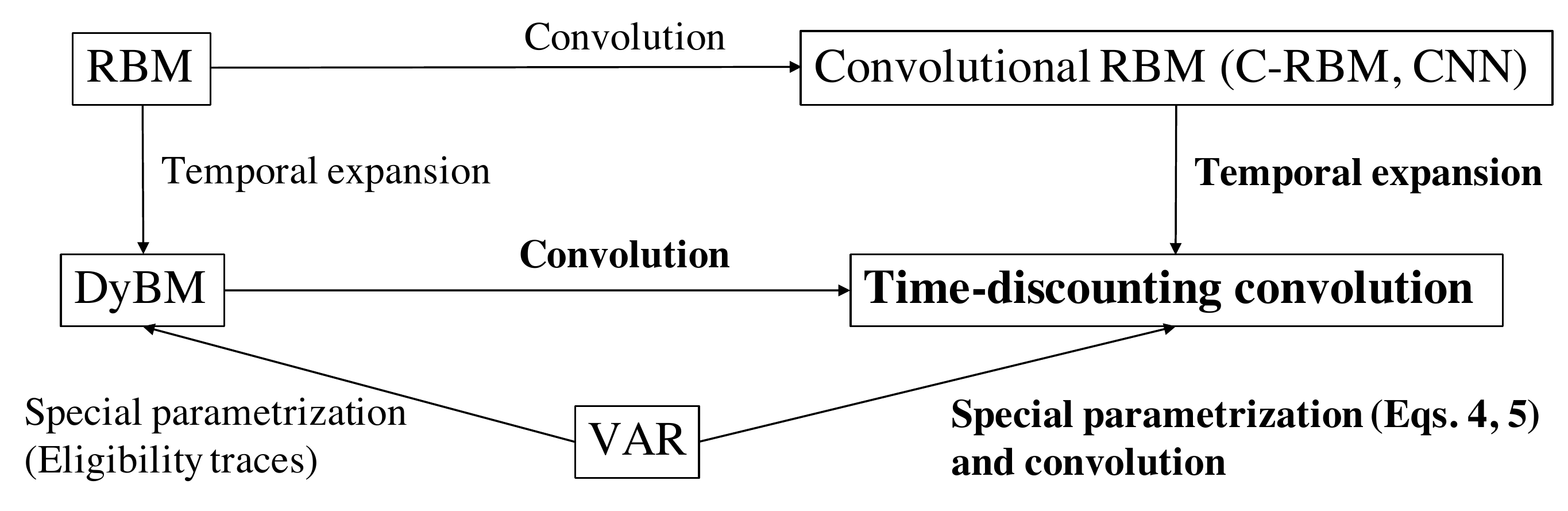}
	\caption{Relationships among RBM, DyBM, C-RBM, VAR, and our model (time-discounting convolution).}
	\label{FigrRelationship}
\end{figure}

DyBM can be considered a temporal expansion of the restricted Boltzmann machine (RBM). By replacing the temporal sequences with hidden variables, the RBM's prediction model for the $j$-th hidden variable is
\begin{equation}
    \label{predict_RBM}
    [f_{\mathrm{RBM}}(\bx,\btheta)]_j = h\Big(\Big\{\sum_{i} x_i W_{k,i,j} - \sum_{i}b_{k,i,j}\Big\}_{k=1}^{K}\Big).
\end{equation}
Its convolutional extension is a convolutional restricted Boltzmann machine (C-RBM). For two-dimensional data, the prediction model for the $(i,j)$-th hidden variable is
\begin{equation}
    \label{predict_CRBM}
    [f_{\mathrm{CRBM}}(\bx,\btheta)]_{i,j} = h\Big(\Big\{\sum_{r,s} W_{k,r,s} x_{i+r,j+s} - b_{k} \Big\}_{k=1}^{K}\Big),
\end{equation}
From Eq.~\eqref{predict_CRBM}, our model can be seen as a temporal expansion of a C-RBM with $1D$-convolution and special parameterizations in Eqs.~\eqref{w_decay} and~\eqref{w_conv} having exponential decay inspired by DyBM. We summarize these relationships in Fig.~\ref{FigrRelationship}.

\section{Experimental Results}
\label{sec:exp}
We assessed the effectiveness of our method in numerical experiments. First, we applied our method to a real-world event sequence with ambiguous timestamps extracted from an EHR. Since our method was designed for general event sequences including ordinary time-series data, we then evaluated its effectiveness for real-world time-series data.

\subsection{Prediction from Real-world Event Sequence with Ambiguous Timestamps}
We evaluated the proposed method using two real-world event sequence datasets, EHRs for patients at a Japanese hospital~\cite{makino2018artificial}. The first dataset included data on $30{,}117$ patients treated for diabetic nephropathy (DN). We constructed a model for predicting progression of DN from stage $1$ to stage $2$ after $180$ days from the latest record in the input EHR (binary classification task). The progression label of the $i$-th input EHR was defined as $y_i \in \{0,1\}$ such that $y_i = 0$ means that the patient remained in stage~$1$ and $y_i = 1$ means that the patient had progressed to stage~$2$. The $i$-th input EHR was represented as a $180$-day sequence of real-valued results of the lab tests, where we represented the sequence as a matrix $\bX_i \in \mathbb{R}^{D \times T}$ for which the horizontal dimension corresponds to the timestamp (time length $T=180$) and the vertical dimension corresponds to the lab tests having $D=25$ attributes, i.e., Albumin, Albuminuria, ALT(GPT), Amylase, AST(GOT), Blood Glucose, Blood Platelet Count, BMI, BUN, CPK, CRP, eGFR, HbA1c, Ht, Hgb, K, Na, RBC, Total Bilirubin, Total Cholesterol, Total Protein, Troponin, Uric acid, WBC count, and $\gamma$-GTP.
The second dataset included data on $36{,}502$ patients treated for cardiovascular disease (CVD). We constructed a model for predicting the occurrence of major cardiovascular events after $180$ days from the latest record in the input EHR (binary classification task). The label of the $i$-th input EHR was defined as $y_i \in \{0,1\}$ such that $y_i = 0$ means that the patient did not experience any of the events and $y_i = 1$ means that the patient had experienced the event. The definition of $\bX_i$ was the same as for the DN case.
For both tasks, following~\cite{katsuki2018risk}, the first $67 \%$ of each dataset was used for training, and the remaining $33 \%$ was used for testing. We standardized the attribute values by subtracting its mean and dividing by its standard deviation in the training data.
\subsubsection{Implementation}
Since we were solving the classification tasks, we used Eq.~\eqref{EqPredictionError_class} for the objective function.
We show the overall structure of the proposed method for the experiments in Fig.~\ref{FigrStructure}. We first applied dynamic pooling to the raw matrix $\bX$. Then, the outputs of the first dynamic pooling were inputted to time-discounting convolution. After that, we applied dynamic pooling again. Finally, we used a fully connected neural network as $h(\bullet)$ in Eq.~\eqref{predict_CDyBM},
$
\sum_{k,d} W^{[f]}_{k,d} \delta(g_{k,d}) + b^{[f]}_j,
$
where $\bW^{[f]}$ and $\bm{b}^{[f]}$ are parameters, $\bg$ is the inputs to $h(\bullet)$ in Eq.~\eqref{predict_CDyBM}, and the function $\delta(\bullet)$ is a rectified linear unit (ReLU)~\cite{nair2010rectified}.
We also used L1-regularization for the hidden units, which are outputs of the second dynamic pooling, in the optimization of Eq.~\eqref{EqPredictionError}. We tuned the regularization parameter $c$ of L1-regularization and hyperparameters of the proposed method using the last $20 \%$ of the training data as validation data. We then trained the model using all of the training data and the tuned parameters. The candidates for $c$ were $\{10^{-2},10^{-1},10^{0}\}$. The hyperparameters candidates were $K \in \{ 4, 8, 16, 24\}$, $\lambda, \mu \in \{ 0.8, 0.85, 0.9, 0.95\}$, $l_0 \in \{ 1, 2, 3, 4, 5,10\}$, and $l \in \{ 1.0, 1.05, 1.1, 1.2\}$. We used four different $T_k$: $1$, $2$, $4$, and the sequence length. We used them in the same proportion in our $K$ feature maps.

\subsubsection{Results}
\begin{figure}[tb]
	\centering
	\includegraphics[width=65mm]{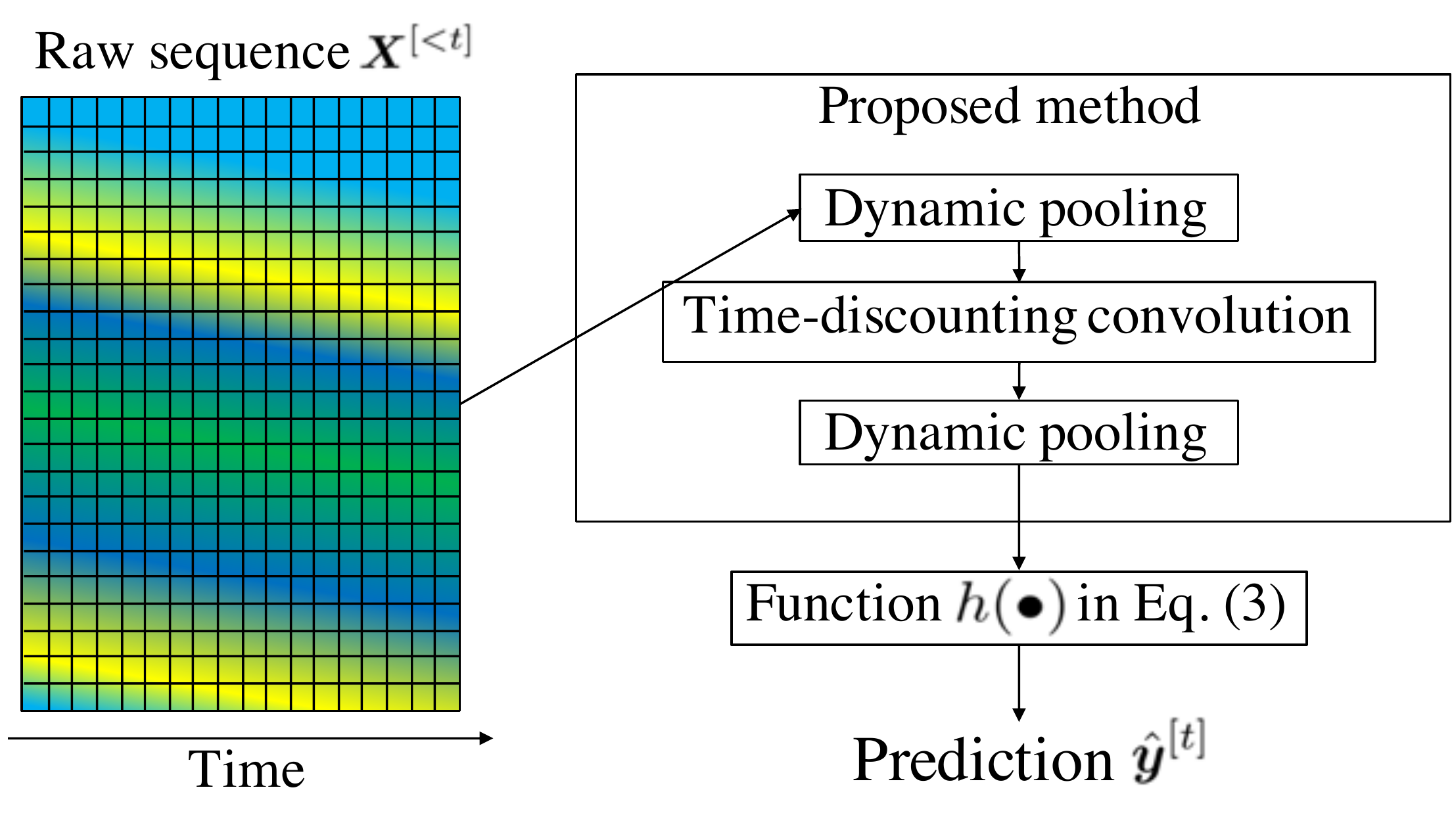}
    \vspace{-5pt}
	\caption{Overall structure of proposed method for experiments.}
	\label{FigrStructure}
\end{figure}
\begin{table}[t]
\caption{Comparison of proposed and baseline methods in terms of AUC (higher is better).}
\vspace{-2pt}
\label{tabPredResults_event}
\centering
\begin{tabular}{ccc}
\toprule
Task & DN & CVD\\
\midrule
DyBM &$0.607$ & $0.620$\\
CNN &$0.647$ & $0.635$\\
CNN w/ dynamic pooling &$0.664$ & $0.636$\\
Proposed w/o dynamic pooling &$0.660$ & $0.635$\\
Proposed w/ dynamic pooling&$\bm{0.674}$ & $\bm{0.647}$\\
\bottomrule
\end{tabular}
\end{table}
\begin{table}[t]
\caption{Comparison of proposed and baseline methods in terms of RMSE (lower is better).}
\vspace{-2pt}
\label{tabPredResults_time}
\centering
\begin{tabular}{ccccc}
\toprule
Task&\multicolumn{2}{c}{Sunspot}&\multicolumn{2}{c}{Price}\\
& Average& Best & Average & Best\\
\midrule
VAR&$0.213$&$0.213$&$0.329$&$0.165$\\
DyBM&$0.0734$&$0.0717$&$\bm{0.0466}$&$0.0302$\\
CNN&$0.0751$&$0.0697$&$0.0845$&$0.0306$\\
Proposed&$\bm{0.0719}$&$\bm{0.0690}$&$0.0489$&$\bm{0.0291}$\\
\bottomrule
\end{tabular}
\end{table}
We used the area under the curve (AUC) as the evaluation metric since the tasks were binary classification. We compared the results against those of two baseline methods, DyBM, a state-of-the-art model for the time-series data, and CNN, a state-of-the-art model for EHR analysis. For fair comparison, we tuned their hyperparameters in the same manner as with the proposed method for each task. For prediction with DyBM, we used the prediction model of DyBM defined in Eq.~\eqref{predict_DyBM} (DyBM). For the prediction with CNN, we used the prediction model by replacing the time-discounting convolution in Fig.~\ref{FigrStructure} with ordinary convolutional layer in Eq.~\eqref{predict_CRBM} (CNN) and that with the two dynamic pooling in Fig.~\ref{FigrStructure} (CNN w/ dynamic pooling). We also present the results of the proposed method without dynamic pooling.
As shown in Table~\ref{tabPredResults_event}, the AUC for the proposed method was better than those for the baselines. This shows that the convolutional structure and our temporal parameterization work well for event sequences with ambiguous timestamps. Moreover, the values for the DN dataset were higher than the $0.64$ reported for a stacked convolutional autoencoder model using the same dataset~\cite{katsuki2018risk}.
The selected hyperparameters for the proposed method were $c=0.01$, $K=8$, $\lambda = 0.95$, $\mu=0.95$, $l_0=4$, and $l=1.05$ for DN, and $c=1.0$, $K=8$, $\lambda = 0.80$, $\mu=0.80$, $l_0=1.0$, and $l=1.05$ for CVD.

\subsection{Prediction from Real-world Time-series Data}
We also evaluated the proposed method using two real-world time-series datasets. The first dataset was a publicly available dataset containing the monthly sunspot number (Sunspot). We constructed a model for predicting the sunspot number for the next month from the obtained sunspot time-series data (autoregression task). The time-series data $\bX_i \in [0,1]^{D \times T}$ had $D=1$ dimension and $T=2,820$ time steps (corresponding to January $1749$ to December $1983$).
The second dataset was a publicly available dataset containing the weekly retail gasoline and diesel prices (Price). We constructed a model for predicting the prices for the following week from the obtained price time-series data (autoregression task). The $\bX_i \in [0,1]^{D \times T}$ had $D=8$ dimensions (corresponding to eight locations in the U.S.) and $T=1,223$ time steps (corresponding to April $5$th, $1993$, to September $5$th, $2016$).
For both tasks, following~\cite{NonlinearDyBM}, the first $67 \%$ of each time series was used for training, and the remaining $33 \%$ was used for testing. We normalized the values of each dataset in such a way that the values in the training data were in $[0,1]$ for each dimension, as in~\cite{BidirectionalDyBM}.

\subsubsection{Implementation}
Since we were solving the autoregression tasks, we used Eq.~\eqref{EqPredictionError_autoreg} for the objective function.
In these tasks, the overall structure of the proposed method and the hyperparameters candidates were the same as in the event-sequence experiment in Section IV-A.

\subsubsection{Results}
We evaluated the methods by using the average test root mean squared error (RMSE) after $1000$ iterations and that of the best case. We compared the results against those of three baseline methods, VAR, DyBM, and CNN. For DyBM and CNN, we used as the same implementation of them in the event-sequence experiment. For VAR, we simply used Eq.~\eqref{predict_VAR} for the prediction function.
As shown in Table~\ref{tabPredResults_time}, the RMSE for the proposed method was comparable to or better than those of the baselines. Moreover, the RMSE values for the proposed method were lower than the $0.0734$~\cite{NonlinearDyBM} and $0.0698$~\cite{BidirectionalDyBM} for the Sunspot data and the $0.0564$~\cite{NonlinearDyBM} and $0.0399$~\cite{BidirectionalDyBM} for the Price data, which were reported as the results from experiments including other DyBM variants and LSTM models. These results indicate that the convolutional structure and our temporal parameterization work well even for ordinary time-series data.
The selected hyperparameters for the proposed method were $c=0.01$, $K=4$, $\lambda = 0.85$, $\mu=0.85$, $l_0=1.0$, and $l=1.0$ for Sunspot, and $c=0.01$, $K=8$, $\lambda = 0.85$, $\mu=0.85$, $l_0=1.0$, and $l=1.0$ for Price.

\section{Conclusion}
We proposed a time-discounting convolution method that can handle time-shift invariance in event sequences and has robustness against the uncertainty in timestamps while maintaining the important capabilities of time-series models. Experimental evaluation demonstrated that the proposed method was comparable to or even better than state-of-the-art methods in several prediction tasks using event sequences with ambiguous timestamps and ordinary time-series data.
The next step in our work is to develop a learning algorithm in an online manner for the proposed method. Actually, we can approximately update the model parameters in an online manner without back propagation through infinite sequences or storing infinite sequences by leveraging dynamic pooling.
Increasing the interpretability of our method is another interesting next step.
\section*{Acknowledgments}
Takayuki Katsuki and Takayuki Osogami were supported in part by JST CREST Grant Number JPMJCR1304, Japan.

\appendix
\subsection{Specific gradients for model parameters}
Here, we show the gradients used in our learning of the model parameters by mini-batch gradient descent in Section III-C.
The gradient of $\mathcal{L}^{[m]}(\btheta)$ with respect to the parameter $U_{k,i,j}$ is
\begin{equation}
    \label{gradient_w_decay}
    \frac{\partial \mathcal{L}^{[m]}(\btheta)}{\partial U_{k,i,j}}=\frac{\partial \mathcal{L}^{[m]}(\btheta)}{\partial \bm{f}(\bX^{[<m]},\btheta)} \bigg[\frac{\partial \bm{f}(\bX^{[<m]},\btheta)}{\partial \bmeta_{j,k}}\bigg]^\top\frac{\partial \bmeta_{j,k}}{\partial U_{k,i,j}},
\end{equation}
where we shorten the partial set of the inputs of the function $h(\bullet)$ on the right-hand side of Eq.~\eqref{predict_CDyBM} related to the $j$-th element of $\bm{f}(\bX^{[<m]},\btheta)$ and the index $k$ in the $K \times T$-dimensional inputs as $\bmeta_{j,k}$ and
\begin{equation}
\frac{\partial \bmeta_{j,k}}{\partial U_{k,i,j}}= \bigg\{ \sum_{\tau_k} \lambda^{d+\tau_k} x^{[m-d-\tau_k]}_i \bigg\}_{d=1}^{T}.
\end{equation}

The gradient of $\mathcal{L}^{[m]}(\btheta)$ with respect to the parameters $V_{k,\tau_k, i, j}$ and $b_k$, $\frac{\partial \mathcal{L}^{[m]}(\btheta)}{\partial V_{k,\tau_k, i, j}}$ and $\frac{\partial \mathcal{L}^{[m]}(\btheta)}{\partial b_k}$, are almost the same to Eq~\eqref{gradient_w_decay} --- they differ only for the gradient of $\bmeta_{k, j}$ with regard to $V_{k,\tau_k, i, j}$ and $b_k$. We show them simply as
\begin{equation}
    \label{gradient_w_conv}
\frac{\partial \bmeta_{j,k}}{\partial V_{k,\tau_k,i,j}}= \bigg\{ \mu^{d} x^{[m-d-\tau_k]}_i \bigg\}_{d=1}^{T},
\end{equation}
\begin{equation}
    \label{gradient_b}
\frac{\partial \bmeta_{j,k}}{\partial b_{k}}= \{ -1 \}_{d=1}^{T}.
\end{equation}

Here, $\frac{\partial \mathcal{L}^{[m]}(\btheta)}{\partial \bmeta_{j,k}}$ for autoregression problems with Eq.~\eqref{EqPredictionError_autoreg} is
\begin{align}
    \label{gradient_h_autoreg}
    \frac{\partial \mathcal{L}^{[m]}(\btheta)}{\partial \bm{f}(\bX^{[<m]},\btheta)}=~& 2\bm{f}(\bX^{[<m]},\btheta) - 2\bX^{[m]},
\end{align}
and that for classification problems with Eq.~\eqref{EqPredictionError_class} is
\begin{align}
    \label{gradient_h_class}
    \frac{\partial \mathcal{L}^{[m]}(\btheta)}{\partial \bm{f}(\bX^{[<m]},\btheta)}=~&
    \frac{1-\by^{[m]}}{1-\bsigma(\bm{f}(\bX^{[<m]},\btheta))}-\frac{\by^{[m]}}{\bsigma(\bm{f}(\bX^{[<m]},\btheta))}.
\end{align}

Through the function $h(\bullet)$ and the gradient $\frac{\partial \bm{f}(\bX^{[<m]},\btheta)}{\partial \bmeta_{j,k}}$, other functions, such as the activation function, and other layers can be applied to our model and the learning algorithm.

\bibliographystyle{IEEEtran}
\bibliography{IEEEabrv,ref.bib}
\end{document}